\documentclass{article} 
\usepackage[letterpaper,textwidth=6.0in,textheight=9.0in]{geometry} 
\usepackage[OT1]{fontenc} 
\usepackage{graphicx} 
\usepackage[table]{xcolor}
\usepackage{caption}
\usepackage{authblk}
\usepackage{subcaption}
\usepackage{rotating}
\usepackage{makecell}
\usepackage{newtxtext,newtxmath} 
\usepackage[np, autolanguage]{numprint}
\usepackage{ctable}
\usepackage{capt-of}
\usepackage{titlesec}
\usepackage{float}
\usepackage{multirow}
\usepackage{longtable}[=v4.13]
\usepackage{tabu}
\usepackage{afterpage}
\usepackage{placeins}
\usepackage{siunitx}
\usepackage{lineno}
\usepackage{ulem}
\usepackage{array}
\newcolumntype{?}{!{\vrule width 1pt}}
\usepackage{authblk}
\usepackage[symbol]{footmisc}
\usepackage{natbib}
\usepackage{bibunits}
\usepackage{etoolbox}
\PassOptionsToPackage{hyphens}{url}\usepackage{hyperref}

\makeatletter
\newcommand*{\newbibstartnumber}[1]{%
  \apptocmd{\thebibliography}{%
    \global\c@NAT@ctr #1\relax
    \addtocounter{NAT@ctr}{-1}%
  }{}{}%
}
\makeatother

\definecolor{LightYellow}{rgb}{1,1,0.88}

\graphicspath{{figures/}}
\setcitestyle{square,comma,numbers,sort}

\newcommand{\RNum}[1]{\uppercase\expandafter{\romannumeral #1\relax}}
\titleformat{\paragraph}
{\normalfont\normalsize\bfseries}{\theparagraph}{1em}{}
\titlespacing*{\paragraph}
{0pt}{3.25ex plus 1ex minus .2ex}{1.5ex plus .2ex}

\title{\rule{\linewidth}{4pt}\\[0.75em]
\textbf{Towards comprehensive cellular characterisation of H\&E slides}\\[0.25em]
\rule{\linewidth}{1pt}}

\author{
Benjamin Adjadj$^{*,\dagger}$,
Pierre-Antoine Bannier$^{*,\dagger}$,
Guillaume Horent$^{*}$,
Sebastien Mandela,
Aurore Lyon,
Kathryn Schutte,
Ulysse Marteau,
Valentin Gaury,
Laura Dumont,
Thomas Mathieu,
MOSAIC consortium,
Reda Belbahri,
Benoît Schmauch,
Eric Durand,
Katharina Von Loga,
Lucie Gillet$^{*}$
}

\affil{Owkin, Paris, France}

\date{}

\begin{document}

\captionsetup[figure]{labelfont={bf},name={Fig.},labelsep=period}
\nprounddigits{2}
\maketitle

\footnotetext[1]{Core team.}
\footnotetext[2]{Equal contributions.}
\begin{bibunit}[unsrt]
\section*{\centering Abstract}

Cell detection, segmentation and classification are essential for analyzing tumor microenvironments (TME) on hematoxylin and eosin (H\&E) slides. Existing methods suffer from poor performance on understudied cell types (rare or not present in public datasets) and limited cross-domain generalization. To address these shortcomings, we introduce HistoPLUS, a state-of-the-art model for cell analysis, trained on a novel curated pan-cancer dataset of 108,722 nuclei covering 13 cell types. In external validation across 4 independent cohorts, HistoPLUS outperforms current state-of-the-art models in detection quality by 5.2\% and overall F1 classification score by 23.7\%, while using 5x fewer parameters. Notably, HistoPLUS unlocks the study of 7 understudied cell types and brings significant improvements on 8 of 13 cell types. Moreover, we show that HistoPLUS robustly transfers to 2 oncology indications unseen during training. To support broader TME biomarker research, we release the model weights and inference code (see Code Availability).

\section{Introduction}
Cell detection, segmentation and classification in histological images form the foundation of modern computational pathology, enabling quantitative analysis of the tumor microenvironment (TME) for applications in patient stratification \cite{ref1,ref2} and personalized medicine \cite{ref3,ref4,ref5}. Beyond cell identification, these techniques reveal spatial relationships between tumor cells, immune cells, and stromal compartments that correlate with disease outcomes \cite{ref6,ref7} and treatment responses \cite{ref8,ref9}. The spatial organization and density of specific cell populations -- such as tumor-associated neutrophils (linked to immunosuppression \cite{ref10}) or cancer-associated fibroblasts (implicated in therapy resistance \cite{ref11,ref12,ref13}) -- have become increasingly important in clinical research and trials \cite{ref14,ref15}. While immunohistochemical methods provide targeted molecular information, H\&E staining remains predominant in clinical practice due to its accessibility, cost-effectiveness, and rich morphological detail \cite{ref16}.
\\
However, extracting comprehensive cellular information from H\&E images presents substantial computational challenges. Cells in tissue sections exhibit marked heterogeneity in morphology, size and staining characteristics, while often being densely packed with indistinct boundaries. The identification of clinically relevant immune cells (e.g., lymphocytes, neutrophils, eosinophils, plasmocytes) and stromal cell subsets (e.g., fibroblasts, smooth muscle cells) requires extensive annotated training data from expert pathologists. Yet, to the best of our knowledge, current public annotated datasets have critical shortcomings: many datasets provide segmentation masks without classification labels \cite{ref17,ref18,ref19}, limiting utility for TME studies; others offer only bounding boxes \cite{ref20}, precluding morphometric analysis. Even datasets with cell-type annotations have limited class granularity \cite{ref21,ref22,ref23}, lack rare cell representation \cite{ref21} or focus on a limited set of cancer types \cite{ref24}. These constraints, coupled with insufficient sample sizes \cite{ref23,ref25}, have hindered the development of robust, generalizable models across diverse tissues and diseases.
\\
\\
Recent pathology foundation models (PFMs) pre-trained via self-supervision \cite{ref26,ref27,ref28} promise to mitigate these limitations by leveraging unlabeled histology data to learn transferable representations, reducing reliance on scarce expert annotations. Models like UNI \cite{ref29}, Phikon \cite{ref30,ref31}, and H-Optimus-0 \cite{ref32} have advanced patch classification and molecular prediction \cite{ref33}, but their applicability to cell detection, segmentation and classification -- particularly for understudied subtypes, i.e.\ non-lymphoid immune cells (eosinophils, neutrophils and macrophages), plasmocytes, stromal cells (smooth muscle, endothelial) and rare events (mitotic figures, apoptotic bodies) -- has not been thoroughly investigated. Despite evaluations of Hibou-Large \cite{ref34}, UNI, Virchow and GigaPath \cite{ref35} on cell-level tasks, no benchmark compares PFMs performance rigorously and exhaustively for simultaneous cell detection, segmentation and classification tasks. This gap is critical, as without systematic benchmarking of PFMs, researchers cannot identify which models maximize performance for these tasks -- leaving potential performance gains unrealized despite their importance for quantifying the TME.
\\
\\
To address these gaps, we present:
\begin{itemize}
    \item A pan-tumor dataset with granular labels for 13 cell types across 6 cancer indications, obtained using an active-learning based annotation pipeline that maximizes label quality and diversity,
    \item A PFM-integrated model for pan-cancer cell detection, segmentation and classification in H\&E, leveraging self-supervised distilled features \cite{ref31,ref32,ref36} within a CellViT architecture that achieves state-of-the-art performances on all nuclei identification, including understudied types.
\end{itemize}

Our model demonstrates superior performance both in cross-validation and on external validation. It paves the way for new analyses of spatial tumor-immune interactions. To facilitate community adoption and further progress in computational pathology, we release our pre-trained CellViT model. This contribution provides the research community with a state-of-the-art cell detection, segmentation and classification model on H\&E slides. It aims to accelerate biomarker discovery, ultimately enhancing our ability to translate histological insights into clinically relevant applications.

\afterpage{%
\begin{figure}[!t]
  \label{fig1}
  \centering

  \begin{subfigure}{\textwidth}
    \label{fig1a}
    \llap{\textbf{a}\hspace{0.6em}}
    \includegraphics[width=\linewidth]{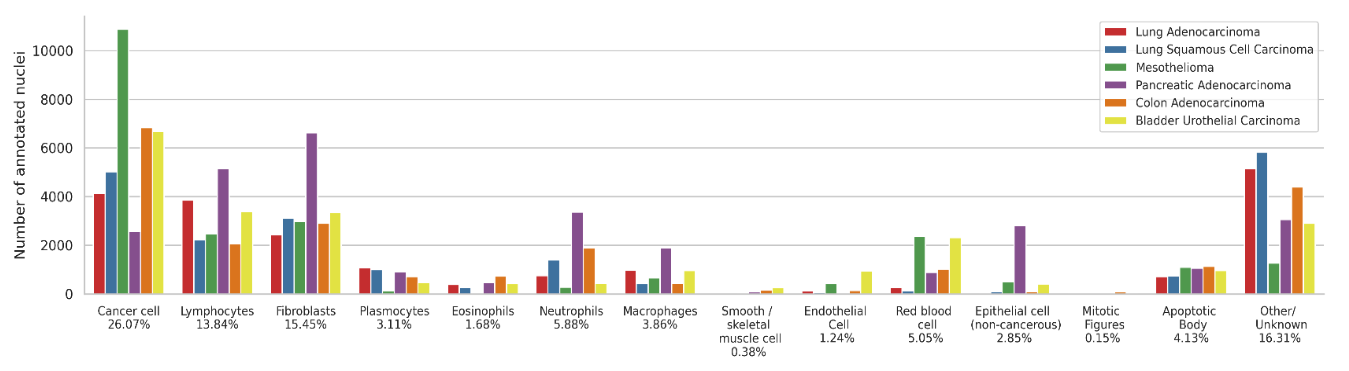}
  \end{subfigure}

  \vspace{0.75\baselineskip}

  \begin{subfigure}{\textwidth}
    \label{fig1b}
    \llap{\textbf{b}\hspace{0.6em}}
    \includegraphics[width=\linewidth]{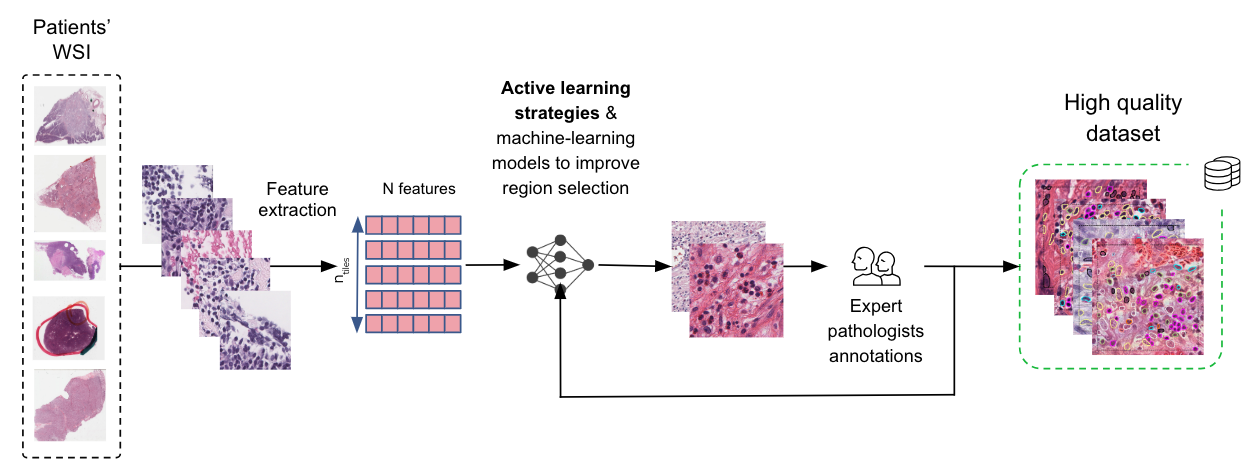}
  \end{subfigure}

  \vspace{0.75\baselineskip}

  \begin{subfigure}{\textwidth}
    \label{fig1c}
    \llap{\textbf{c}\hspace{0.6em}}
    \includegraphics[width=\linewidth]{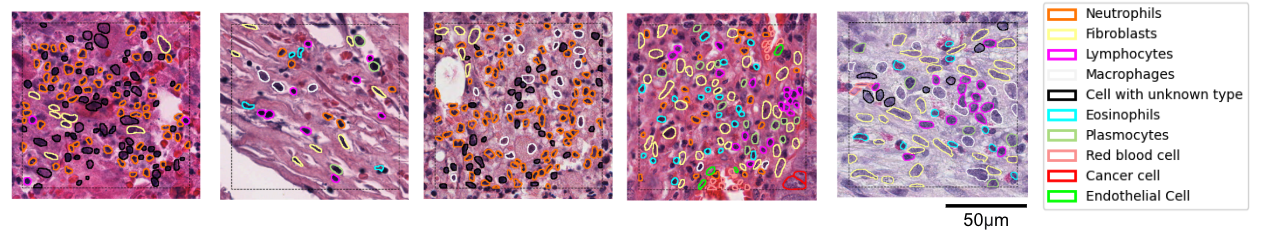}
  \end{subfigure}

  \caption{\textbf{Overview of our training dataset.}
  \textbf{a)} Distribution of annotated nuclei across cell types and cancer indications. Barplot showing the number of annotated nuclei per cell type across six cancer types. The active learning pipeline enhanced the enrichment of our dataset in understudied cell types on H\&E.
  \textbf{b)} Candidate whole-slide images (WSIs) are divided into tiles, and features are extracted from each tile using Phikon, a pathology foundation model. To prioritize informative and diverse tiles, we apply K-means clustering based on features as well as multiple active learning strategies: rare cell type enrichment using cell type predictors (MLPs)  and uncertainty estimation. The set of tiles is submitted to a pool of expert pathologists for cell type annotation.
  \textbf{c)} Examples of 448$\times$448 images at 40$\times$ magnification selected through active learning, annotated independently by multiple expert pathologists, and consolidated using consensus guidelines.}
  \label{fig:training_dataset}
  \vspace{1em}
\end{figure}
}

\section{Results}\label{sec:res}

  \subsection{Pan-cancer nuclei dataset enriched in understudied types enables granular study of TME}%
  \label{subsec:Results/paragraph_1}%
  We introduce HistoTRAIN, a pan-cancer nuclei dataset designed to address critical gaps in cellular diversity, annotation quality, and clinical applicability. The dataset comprises 108,722 nuclei segmentations from 739 H\&E whole-slide images (WSI), across six cancer types (bladder urothelial carcinoma, colon adenocarcinoma, lung adenocarcinoma, lung squamous cell carcinoma, mesothelioma and pancreatic adenocarcinoma) and covering 13 distinct cell types (Figure 1a). This dataset collection was done in 2 steps. First, point annotations of nuclei centroids together with their respective classes were entirely made by expert pathologists. Second, we leveraged NuClick \cite{ref37}, a deep learning model specifically trained for inferring nuclei segmentations from point annotations in nuclei images, to obtain precise segmentations for each annotated nucleus. Leveraging NuClick for nuclei contours drastically reduces the variability in manual boundary delineation and accelerates the annotation process.
\\
\\
To systematically prioritize rare and relevant cell types, we developed an active learning pipeline (Figure 1b) that iteratively identifies tissue regions with low model confidence and high probability of understudied cell types presence (see Methods). This approach increased the representation of understudied cell types compared to existing public datasets (see Table 1). The dataset is enriched with nuclei segmentations of non-lymphoid immune cells (eosinophils, neutrophils and macrophages), plasmocytes, stromal cells (smooth muscle, endothelial) and rare events (mitotic figures, apoptotic bodies). Epithelial cells (non-cancerous) and red blood cells were also included to improve morphological context. A detailed distribution of each cell type per cancer indication is presented in Figure 1a. In total, our dataset comprises 1,415 images of size 448x448 pixels (at a 40x magnification; Figure 1c).

\begin{table}[h!]
\label{table:table_1}
\centering
\begin{tabular}{l*{5}{c}}
\toprule
\textbf{Cell types} & \textbf{NuCLS} & \textbf{CoNSeP} & \textbf{Lizard} & \textbf{PanNuke} & \textbf{HistoTRAIN (Ours)} \\ 
\midrule
Neutrophils        & 0.09\% & - & 0.97\% & - & \textbf{5.88\%} \\
Eosinophils        & 0.01\% & - & 0.73\% & - & \textbf{1.86\%} \\
Macrophages        & 2.69\% & - & - & - & \textbf{5.29\%} \\
Mitotic Figures    & - & - & - & - & \textbf{0.21\%} \\
Apoptotic Bodies   & - & - & - & 1.53\% & \textbf{5.33\%} \\
Endothelial Cells  & - & - & - & - & \textbf{2.02\%} \\
Smooth Muscle Cells& - & - & - & - & \textbf{1.90\%} \\
\bottomrule
\end{tabular}
\caption{\textbf{Comparison of the distribution of understudied cell populations in our dataset and existing public datasets.} 
Values represent the proportion (\%) of each cell type among all annotated nuclei. “-” indicates that the corresponding nuclei type is not present in the dataset.}
\end{table}

\afterpage{%
\begin{figure}[!t]
  \label{fig2}
  \centering

  \begin{subfigure}{\textwidth}
    \label{fig2b}
    \includegraphics[width=\linewidth]{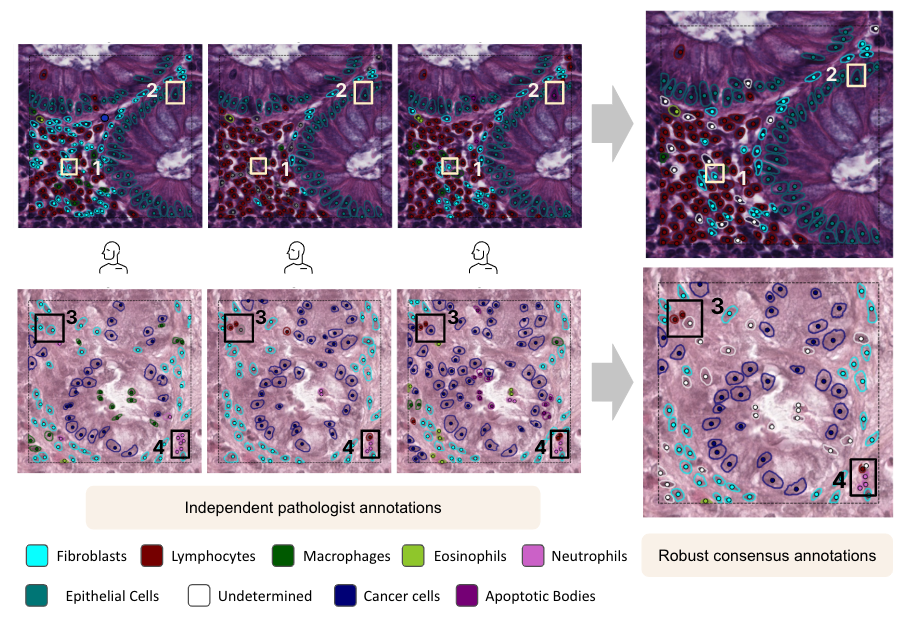}
  \end{subfigure}

  \caption{\textbf{Robust external validation sets.}
  Example of consensus annotations derived from regions annotated independently by 3 expert-pathologists. Disagreement may arise from distinct nuclei chosen to be annotated vs. skipped [1],  distinct cell types being attributed [2], or a mix of both of these reasons [3], [4].}
  \label{fig:robust_validation}
  \label{fig:robust_validation}
  \vspace{1em} 
\end{figure}
}

\afterpage{%
\begin{figure}[!tp]
  \label{fig3}
  \centering

  \begin{subfigure}{\textwidth}
    \label{fig3a}
    \llap{\textbf{a}\hspace{0.6em}}
    \includegraphics[width=\linewidth]{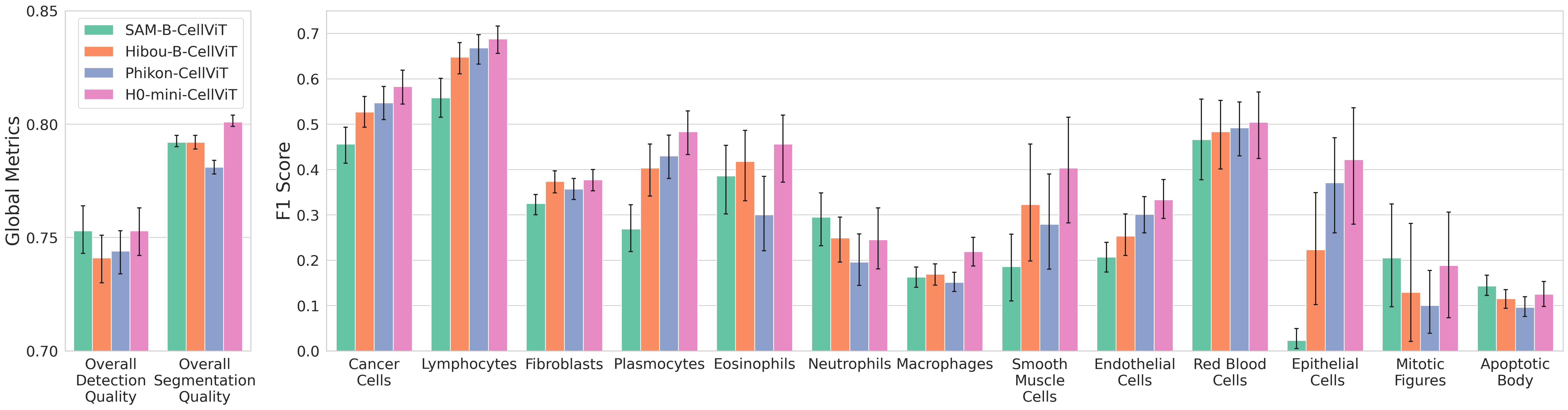}
  \end{subfigure}

  \vspace{0.65\baselineskip}

  \begin{subfigure}{\textwidth}
    \label{fig3b}
    \llap{\textbf{b}\hspace{0.6em}}
    \includegraphics[width=\linewidth]{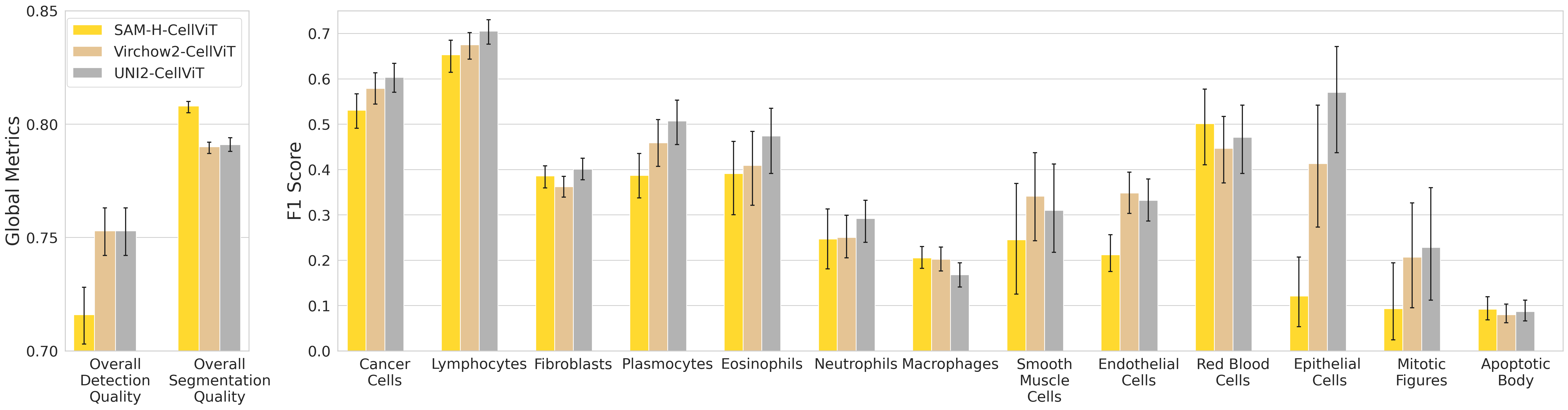}
  \end{subfigure}

  \vspace{0.65\baselineskip}

  \begin{subfigure}{\textwidth}
    \label{fig3c}
    \llap{\textbf{c}\hspace{0.6em}}
    \includegraphics[width=\linewidth]{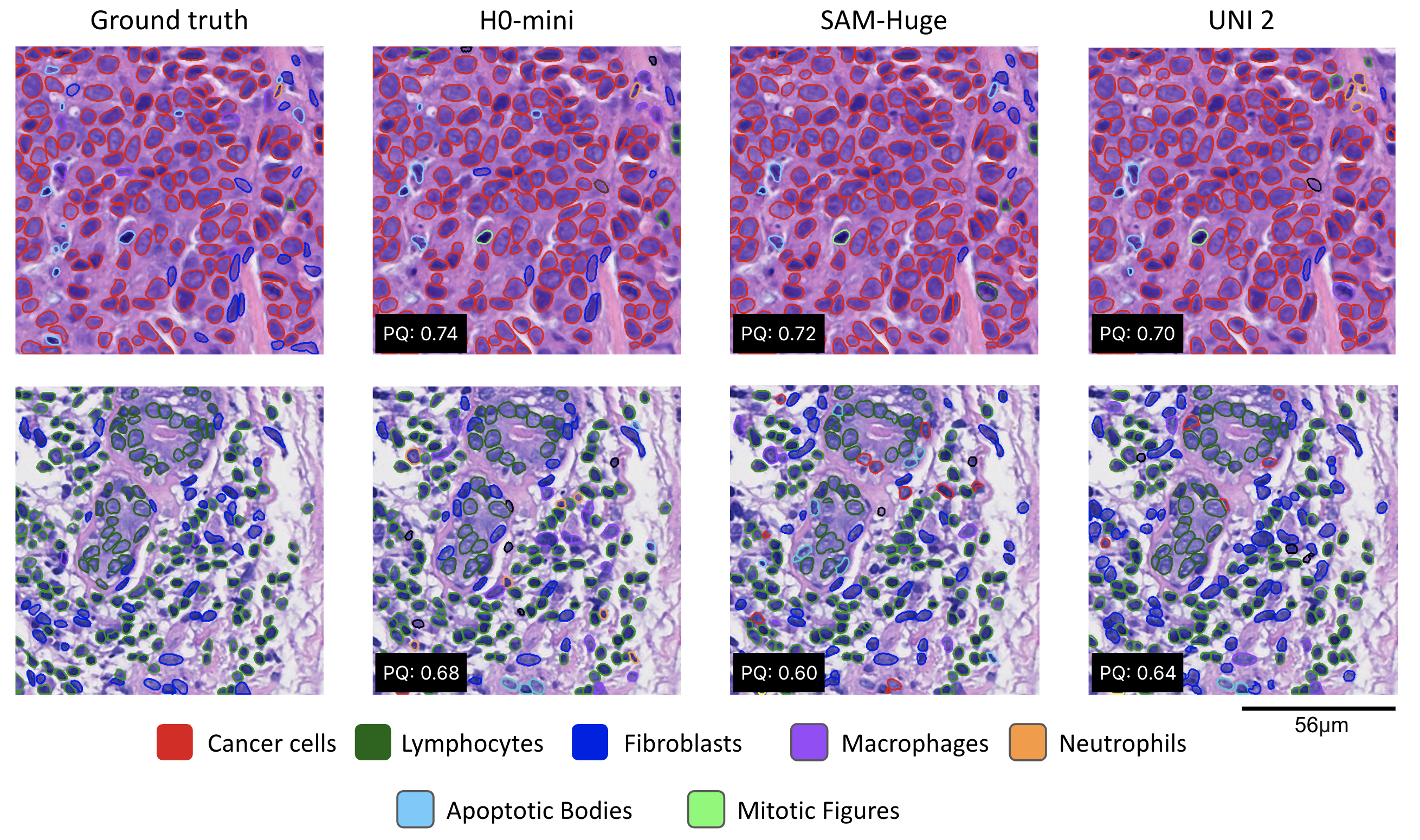}
  \end{subfigure}

  \caption{\textbf{Evaluation of pathology foundation models with the CellViT architecture.}
  \textbf{a)} Performance of CellViT with vision transformer base encoders on external validation. We evaluated the performance of different vision transformer base encoders integrated with CellViT for cell detection and segmentation in external validation. The barplots present the median detection quality, segmentation quality and F1 scores for each encoder, estimated using 1{,}000 bootstrap iterations. Whiskers indicate the 95\% confidence intervals for each metric.
  \textbf{b)} Barplots analogous to (a), showing the performance of vision transformer huge encoders integrated with CellViT on external validation.
  \textbf{c)} Representative examples of cell detection, segmentation and classification for 448$\times$448 images at 40$\times$ magnification (effective field of view is 112$\times$112$\,$\textmu m) by CellViT with H0-mini, SAM-Huge and UNI2. The image-level panoptic quality (PQ) is shown on each image.}
  \label{fig:fm_eval}
  \vspace{1em} 
\end{figure}
}

\afterpage{%
  \begin{figure}[!t] 
    \label{fig4}
    \centering
    \begin{subfigure}{\textwidth}
        \llap{\textbf{a}\hspace{0.6em}}
        \includegraphics[width=\linewidth]{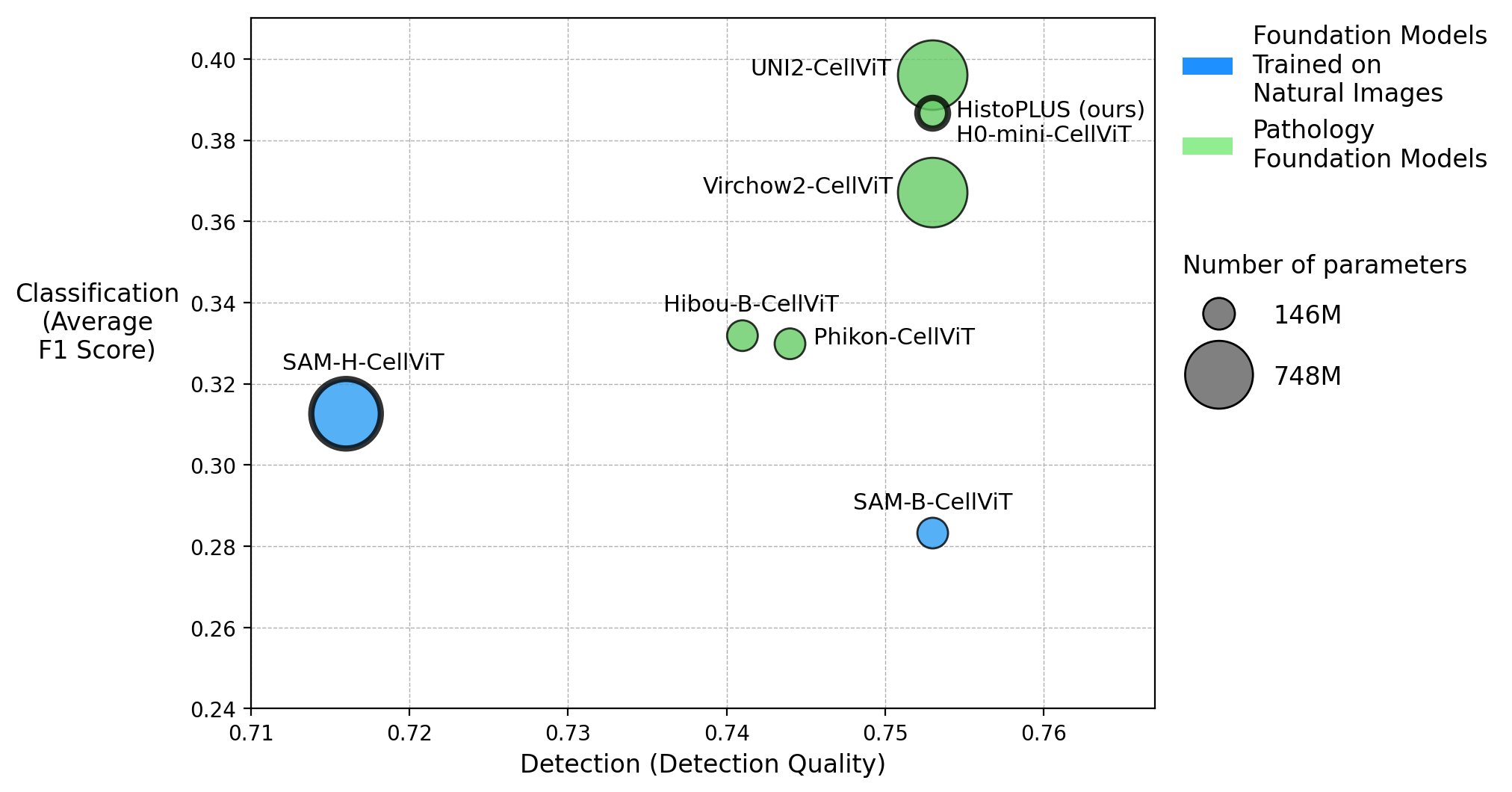}
    \end{subfigure}
    
    \vspace{0.75\baselineskip}

    \begin{subfigure}{\textwidth}
        \llap{\textbf{b}\hspace{0.6em}}
        \includegraphics[width=\linewidth]{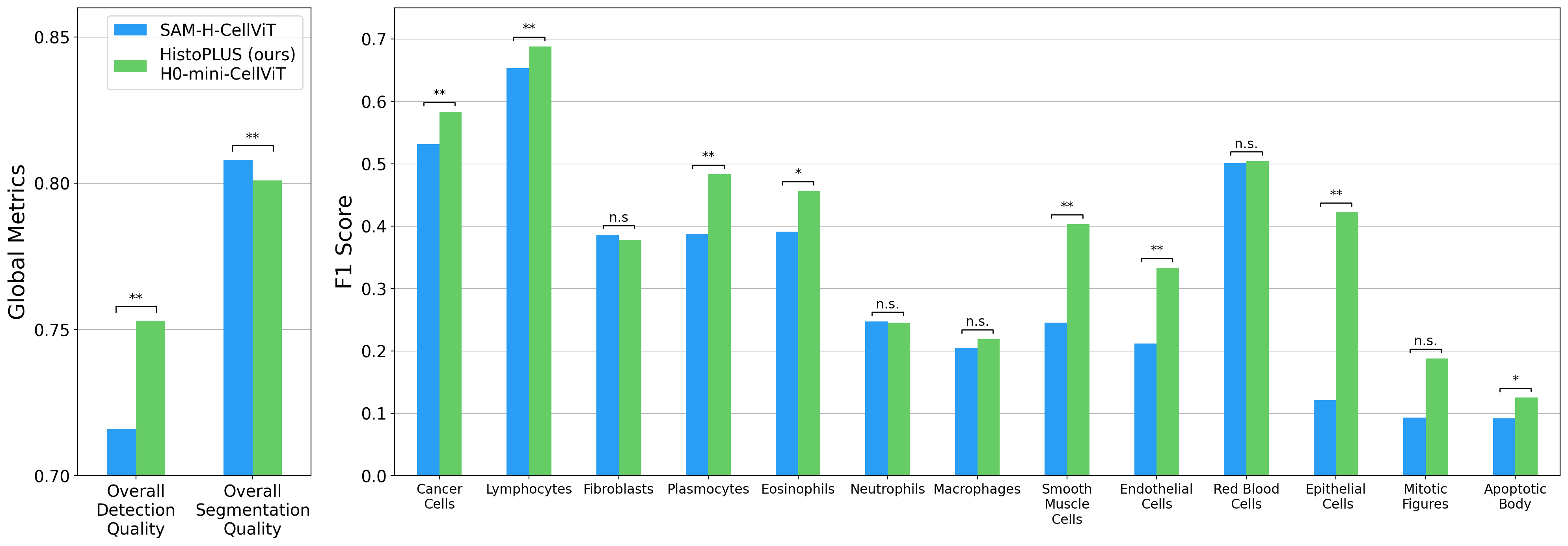}
    \end{subfigure}
    
    \caption{\textbf{HistoPLUS outperforms the current state-of-the-art in detection and classification while reducing model complexity.}
    \textbf{a)} The comparison of CellViT models with different backbones based on detection quality, classification performance and model size confirms the superiority of pathology-specific encoders on external set. Scatter plot showing detection quality (x-axis) and the average F1 score across cell types (y-axis) for CellViT models with various encoders. Each bubble represents a model, with size indicating the number of parameters. Pathology-specific foundation models are shown in green; general-purpose models in blue. The CellViT model with the H0-mini encoder emerges as the best trade-off, combining strong detection and classification performance with a compact architecture. Models highlighted in bold are those compared in panel b).
    \textbf{b)} Head-to-head comparison between CellViT with SAM-H (state-of-the-art) and our model HistoPLUS – CellViT with the H0-mini encoder – on our external test set. The left panel compares detection and segmentation quality, while the right panel shows per-class classification F1 scores. Performances are computed via bootstrapping with 1,000 iterations. Statistical significance is indicated as follows: $^{*}P<0.05,\;^{**}P<10^{-3}$; non-significant differences are labeled "n.s.". Double-sided p-values were computed using bootstrap resampling.}
    \label{fig4}
  \end{figure}
}

  \subsection{Robust external validation sets are built using a consensus of expert pathologists}%
  \label{subsec:Results/paragraph_2}%
  Deriving a reliable ground truth for cell detection, segmentation and classification tasks on H\&E images is challenging due to substantial inter-annotator variability, even among expert pathologists \cite{ref38}. Key sources of variation include cell type attribution and annotation completeness. To address these challenges, we randomly assigned 3 out of 9 pathologists to annotate each tile and used a consensus framework to create curated, reliable external validation sets.
\\
\\
Our consensus framework operates in four sequential steps. First, 2 to 3 pathologists independently annotate nuclei centroids with point annotations on a given set of tiles (number of total annotations prior to consensus = 212,992 nuclei). Second, we infer Nuclick to expand the points to nuclei segmentations. Third, we establish correspondence between annotations from different pathologists by matching segmentations with an Intersection over Union (IoU) above 0.4, retaining only nuclei identified by at least two pathologists to ensure reliability. Finally, for each matched nucleus, we compute a consensus centroid by averaging all corresponding centroid annotations (Supplementary Figure 1). The cell type is determined through majority voting. It is fully determined if more than half of annotators agree on a class. Otherwise, all classes are kept and classification metrics are adjusted accordingly (see Methods). The final consensus segmentations are generated by applying NuClick to the computed consensus centroids, yielding high-quality nuclear boundaries paired with robust cell type classifications. 
\\
\\
We applied this methodology to derive HistoVAL (Figure 2, Supplementary Figure 2), a validation dataset which contains 530 regions of size \qty{112}{\micro\metre} x \qty{112}{\micro\metre} acquired at 40x magnification, selected from 248 distinct slides, covering the same 13 cell types. Regions are selected either randomly or to maximize the presence of understudied cell types. These regions cover a total of 6 cohorts originating from the MOSAIC \cite{ref39} dataset, spanning four cancer types present in the training set (bladder cancer, mesothelioma, lung adenocarcinoma and lung squamous cell carcinoma) and two unseen cancer types (ovarian, breast) to assess transferability on external unseen indications. HistoVAL represents a total of 69,108 consensus-driven nuclei.
  \subsection{Pathology foundation models improve cell classification}%
  \label{subsec:Results/paragraph_3}%
  To evaluate the impact of pathology-specific pretraining on cell detection, segmentation and classification, we integrated PFMs as encoders within the CellViT \cite{ref40} architecture and compared them with the general-purpose Segment Anything Model \cite{ref41} (SAM) using 3-fold cross-validation stratified by cell type on HistoTRAIN. To further evaluate the performance of each model to generalize, we performed external validation on HistoVAL. For cell detection and segmentation, we used the detection quality (DQ) and segmentation quality (SQ) metrics. For cell classification we computed F1 scores for each cell type (see Evaluation metrics for detection, segmentation and classification in Methods).
\\
\\
To compare equal-sized encoders, we evaluated PFMs built on the vision transformer \cite{ref42} (ViT) base architecture against the general-purpose encoder SAM-Base \cite{ref41} (SAM-B). Domain-specific pretraining consistently improved performance over general-purpose models in cross-validation (Supplementary Table 1,2). These findings are also confirmed in external validation (n=530): Phikon \cite{ref31}, Hibou-B \cite{ref34} and H0-mini \cite{ref36} showed a respective relative improvement of 16.5\%, 17.2\% and 36.5\% in their average F1 score compared to their counterpart SAM-B. For example, H0-mini achieved an F1 score of 0.583 [0.544 - 0.619] on tumor cells compared to SAM-B's 0.456 [0.414 - 0.493]. For understudied cell types, the improvements were even more substantial and statistically significant, showing H0-mini achieved F1 scores of 0.483 [0.433 - 0.529] vs. 0.269 [0.219 - 0.322] for plasmocytes ($P< 0.001$) and 0.403 [0.282 - 0.515] vs. 0.186 [0.110 - 0.257] for smooth muscle cells ($P < 0.001$) (Figure 3a; Supplementary Table 4). Similar results are found as well for large encoders: Virchow2 \cite{ref43} and UNI2 \cite{ref29} showed a respective improvement of 17.4\% and 26.7\% compared to SAM-H \cite{ref41}.
\\
\\
Despite considerable increases in model complexity, larger architectures showed limited performance gains while incurring significant computational costs. ViT-Huge models (UNI2 \cite{ref29}; Virchow 2 \cite{ref43}; 632M parameters) achieved only marginal PQ improvements over the compact H0-mini \cite{ref36} model (86M parameters) in cross-validation (Supplementary Table 1). External validation revealed similarly modest or no improvements, with UNI2 \cite{ref29} and Virchow 2 \cite{ref43} both achieving DQ scores of 0.753 [0.742 - 0.763], compared to H0-mini \cite{ref36} 0.753 [0.742 - 0.763] (Supplementary Table 3; Figure 3b). For cell classification, Virchow 2 \cite{ref43} and UNI2 \cite{ref29} achieved a F1 score on cancer cells of 0.579 [0.544 - 0.613] and 0.603 [0.570 - 0.634], respectively, in-line with H0-mini’s 0.583 [0.544 - 0.619] (Supplementary Table 4; Figure 3b; Figure 3c). Similar patterns were observed across individual cell types, including rare classes where larger models failed to substantially outperform H0-mini \cite{ref36} (Supplementary Table 4). These marginal performance gains came at significant computational cost: 5.1x more parameters and 2.1x times longer inference time (Figure 4a; Supplementary Table 5). Besides, segmentation quality remained consistent across all model sizes ($\Delta_{\text{SQ}}$ $\le$ 1.5), confirming that the primary benefit of domain-specific pretraining lies in classification accuracy rather than boundary delineation.
\\
\\
Taken together, these results motivated the selection of HistoPLUS, a compact yet high-performing configuration combining the CellViT \cite{ref40} architecture with the H0-mini \cite{ref36} backbone, as our reference model. HistoPLUS offers an optimal balance between accuracy and efficiency, with 5× fewer parameters than CellViT SAM-H \cite{ref40} while matching its detection quality and exceeding its classification performance, particularly for rare and clinically relevant cell types (Figure 4b). This balance enables accurate, fine-grained tumor microenvironment profiling at substantially reduced computational cost, making HistoPLUS a practical choice for large-scale deployment.%

  \subsection{HistoPLUS demonstrates robust performance on unseen oncology indications}%
  \label{subsec:Results/paragraph_4}%
  To assess the generalizability of HistoPLUS beyond the indications present in the training set (Supplementary Tables 6 and 7 for performances per indication), we included in our validation datasets samples from 2 external indications: breast cancer (n=90) and ovarian cancer (n=50). HistoPLUS maintained robust detection and segmentation capabilities across both tissue types, achieving detection quality scores of 0.836 [0.819 - 0.852] for breast and 0.805 [0.782 - 0.825] for ovarian samples, with corresponding segmentation quality scores of 0.801 [0.796 - 0.807 and 0.803 [0.795 - 0.810], respectively (Table 2; Supplementary Table 6). The model demonstrated particularly strong performance in classifying lymphocytes across both indications, with F1 scores of 0.799 [0.757 - 0.829] on breast and 0.639 [0.585 - 0.676] on ovarian cases (Supplementary Table 7). It also displayed promising results on cancer cells, with F1 scores of 0.682 [0.594 - 0.751] for ovarian and 0.475 [0.302 - 0.620] for breast cases.

\begin{center}
\resizebox{\textwidth}{!}{%
  \begingroup
    \let\table\relax \let\endtable\relax 
    \label{table:table_2}
\centering
\setlength{\tabcolsep}{5pt} 
\renewcommand{\arraystretch}{1.4} 
\begin{tabular}{l*{6}{c}}
\toprule
\textbf{Cohort} & \textbf{Detection Quality} & \textbf{Segmentation Quality} & \textbf{Cancer Cells} & \textbf{Lymphocytes} & \textbf{Plasmocytes} & \textbf{Eosinophils} \\
\midrule

\textit{Ref.: External,} & 0.725 & 0.801 & 0.582 & 0.568 & 0.510 & 0.464 \\

\textit{seen indications} & [0.713 - 0.738] & [0.798 - 0.804] & [0.539 - 0.620] & [0.534 - 0.596] & [0.453 - 0.557] & [0.373 - 0.535] \\

Ovarian (external; & 0.805 & 0.803 & 0.682 & 0.639 & 0.330 & 0.250 \\

unseen indication) & [0.782 - 0.825] & [0.795 - 0.810] & [0.594 - 0.751] & [0.585 - 0.676] & [0.183 - 0.430] & [0.000 - 0.889] \\
 
Breast (external; & 0.836 & 0.801 & 0.475 & 0.799 & 0.296 & 0.429 \\
unseen indication) & [0.819 - 0.852] & [0.796 - 0.807] & [0.302 - 0.620] & [0.757 - 0.829] & [0.120 - 0.380] & [0.100 - 0.727] \\
\bottomrule
\end{tabular}

  \endgroup
}
\captionof{table}{\textbf{Zero-short generalizability of HistoPLUS model on external cohorts.} 
HistoPLUS is validated on different sets of indications: indications that have been seen at training time (validation cohorts for these indications are distinct from cohorts included at training time) and indications that have not been seen at training time. A comprehensive breakdown can be found in Supplementary Tables 8 and 9.}

\label{tab:zeroshot}
\end{center}

\section{Discussion\label{sec:disc}}
We present HistoPLUS, a model that advances the state-of-the-art of simultaneous cell detection, segmentation and classification. Our results demonstrate that pathology-specific foundation models, and diverse training data together significantly increase performance without requiring overly large models, which is crucial for clinical applications.
\\
\\
The performance gains of HistoPLUS reflect two complementary advances. First, our curated training dataset addresses a critical limitation in existing public resources: the underrepresentation of some clinically important cell types. By employing active learning to systematically enrich our dataset with challenging examples across six cancer types, we achieved more robust classification of infrequent populations such as eosinophils, neutrophils and endothelial cells. Second, the integration of pathology foundation models as encoders provides substantial improvements over general-purpose vision models, with domain-specific pretraining proving particularly beneficial for rarer cell type classification. 
\\
\\
These technical advances open new avenues for H\&E-based biomarker discovery beyond traditional density-based and morphometric features. The ability to accurately quantify diverse cell populations at scale enables the development of sophisticated cellular composition signatures and spatial relationships that could complement or potentially replace more expensive molecular assays. For example, precise quantification of tumor-infiltrating immune cell subtypes, including non-lymphoid populations like neutrophils or eosinophils, may provide insights into inflammatory processes and treatment response prediction. Furthermore, when integrated with tissue-level spatial analysis models, HistoPLUS could enable the characterization of cellular microenvironments–such as neutrophil density in juxtatumoral regions or plasmocyte distribution relative to tumor boundaries–potentially revealing spatial biomarkers invisible to current analysis approaches.
\\
\\
While our cross-indication validation demonstrates promising generalizability, several limitations warrant consideration. The model’s performance on cell types across tissue contexts remains to be fully characterized, and validation on specific cancer subtypes, additional cancer types and non-neoplastic conditions will be essential for broader clinical adoption. Additionally, the computational requirements, while reduced compared to larger foundation models, may still pose challenges for resource-constrained settings. Detecting and classifying all nuclei on a whole-slide image of a surgical resection typically requires 20 to 40 minutes on a single Tesla T4 GPU. Future work should focus on further model compression techniques and the development of uncertainty quantification methods to enhance clinical reliability. Nevertheless, HistoPLUS represents a significant step toward scalable, comprehensive cellular analysis of routine histopathology, with the potential to uncover new H\&E-based biomarkers and more accurate therapeutic stratification.

\section{Methods\label{sec:methods}}
\subsection*{Data collection for training set}

To generate our dataset, we relied on TCGA datasets of Colon adenocarcinoma (COAD), Lung adenocarcinoma (LUAD), Lung squamous cell carcinoma (LUSC), Mesothelioma (MESO), Bladder Urothelial Carcinoma (BLCA), Ovarian serous cystadenocarcinoma (OV), Pancreatic adenocarcinoma (PAAD) and Breast invasive carcinoma (BRCA). As a first step, Whole Slide Images (WSI) are divided into smaller, non overlapping areas, referred to as tiles. Each tile represents a tissue portion of \qty{112}{\micro\meter} x \qty{112}{\micro\meter} either at 40x when available (n=1,322), or at 20x (n=93). We leveraged pretrained foundation models \cite{ref31} to extract tile-level features. We selected an equal number of tiles for each indication, using our active learning pipeline to favorise diverse and informative tiles within cohorts. Selected tiles were then uploaded to Cytomine \cite{ref44}, a web-based interface allowing board-certified pathologists to point nuclei centroids, and select the corresponding cell type amongst a predefined ontology. A “Cell with unknown type” class was also available if pathologists were not fully confident on their guess. Annotations were then pulled from Cytomine for preprocessing.

\subsection*{Data collection for validation sets}

In parallel with collecting training annotations, we collected annotations for validating our final model on external cohorts, relying on MOSAIC datasets. Detailed methods associated with the MOSAIC data are presented in \cite{ref39}. Within available datasets n=248 slides were selected randomly. Within each selected slide, 3 tiles are picked for annotation by random selection, by maximizing the probability of presence of eosinophils and by maximizing the presence of neutrophils, yielding a $1/3$ split for each selection method. Presence of neutrophils and eosinophils is assessed based on the predictions of an ensemble of 3 Multi-Layered-Perceptron (MLP) trained to predict presence of nuclei types at the tile level. These MLPs consist of a single layer MLP with an input layer of size 2048 (taking as input MoCo‑WideResNet \cite{ref45} extracted features) and an output of size 1, with a Sigmoid activation, predicting the presence of at least one cell of the corresponding cell type within the tile. These MLPs were pretrained on 3 datasets (TCGA LUAD, LUSC, COAD) where 636 tiles were annotated (307 tiles were annotated positive for the presence of eosinophils, and 329 for the presence of neutrophils). We also randomly sample the same number of negative examples from our dataset to obtain a balanced representation in the training set of each predictor. This methodology ensures a decent representation of important understudied nuclei types (Supplementary Figure 1).

\subsection*{Building a diverse dataset with active learning}

Using the tiles presented in the Data collection paragraph, we chose regions of interest using A) clustering methods to ensure tissue diversity and B) active learning-based strategies to maximize the representation of areas with rare cells or ambiguous cells. As the inference of large cell detection, segmentation and classification is computationally expensive, we used tile-level classifiers performances (simple predictors estimating the presence/absence of a given cell type within a tile) as proxies for estimating the classification performance of cell detection, segmentation and classification models. It allowed us to leverage active learning strategies (iteratively refine classifiers thanks to newly collected annotations) without the compute and time cost of fitting large cell detection, segmentation and classification models at each iteration. We defined 3 different active learning methods to select tiles within candidate datasets; our goal was to not only generate tiles for annotation, but also to evaluate the performance of tile selection strategies and the added value of rare cell enrichment and uncertainty sampling.
\\
\\
The first method, used as baseline, consisted of feature-based clustering: given a pool of S slides, we performed K-means clustering of the tiles for each slide based on their features extracted using Phikon \cite{ref46}. We sampled randomly one tile from each cluster, yielding K*S tiles for the dataset. We then randomly sampled the desired number of tiles N0 from this pool. For our experiments, we set K = 20 as suggested in the recent literature \cite{ref47}, to ensure homogeneous, diverse tile clusters across the slide.
\\
\\
Our second method was designed to enrich our dataset in under-represented cell types, such as: Apoptotic Bodies, Mitotic Figures, Eosinophils, Endothelial cells, Epithelial cells, Macrophages, and Neutrophils. For every phenotype we trained lightweight MLPs that map either Phikon or MoCo‑WideResNet \cite{ref45} features to a probability of presence, using annotated tiles from TCGA COAD, LUAD and LUSC cohorts and all tiles annotated from the previous iterations (resulting in 341 tiles for the first iteration, and 1782 tiles after all iterations). We performed cross-validation on cohort-stratified folds with nested inner-folds, features are z‑scored with a conditional scaler (global, slide‑wise or cohort‑wise) to cancel slide/cohort biases. Twelve models (2 network depths $\times$ 2 feature sets $\times$ 3 scalers) were compared per cell type and the best one, estimated using cross‑validation balanced accuracy, was refitted on the entire dataset for inference. We scored all available tiles and kept the top two highest scored tiles, per slide and per cell type. The union of these high‑confidence tiles were merged with the diversity pool before the final sampling step (random sampling from the pool), ensuring that rarer cell types are oversampled while the slide‑level weighting keeps the overall distribution balanced.
\\
\\
To further prioritise informative examples, we computed epistemic uncertainty on the rare‑cell predictions in our third arm. Each phenotype model was run as an ensemble; from the per‑tile logits we derived entropy, variation ratio and BALD scores \cite{ref48}. Among tiles whose predicted probability were above 0.4, the ones whose BALD score (high uncertainty) exceeded the 90th percentile on their slide were retained. For those uncertain tiles, we performed K‑means on each slide (between 5 and 20 clusters depending on remaining tile count) and took one representative per cluster. This yielded a compact uncertainty‑aware pool. During the final draw we sampled 50 tiles per indication, proportional to the number of uncertain tiles a slide contributes but clipping slide weights to the 0.5–2.0 range to avoid over‑ or under‑representation.
\\
\\
This resulted in a dataset that (i) covered the morphological space of every slide, (ii) deliberately enriched rare phenotypes, and (iii) focused annotation effort on regions where the current models disagree most, accelerating both cell‑type and segmentation learning in later active‑learning rounds.

\subsection*{Cell boundary extension with Nuclick}

Handmade segmentations are time-consuming to perform and highly subject to pathologist inter-variability, which prevents building a large, high-quality set of annotations for model training and testing, across multiple indications. In order to speed up and robustify the annotation collection process, we asked pathologists to solely annotate the centroid of each nucleus with a point annotation along with the corresponding cell type, and processed these pre-annotations using Nuclick \cite{ref49} to retrieve nuclei segmentations from pathologist point annotations. Nuclick is a CNN-based prompted segmentation network which takes as input a map of annotated nuclei centroids within a tile and returns the corresponding nuclei segmentations as an instance map. We used the vanilla Nuclick U-Net architecture as described in the official Github repository (\url{https://github.com/mostafajahanifar/nuclick_torch/}), and trained our model on ConSeP \cite{ref22} and PCNS \cite{ref49} datasets, which regroups annotated data from 14 cancer indications. For each annotated nucleus in the training set, we extracted a 128x128 patch centered on the cell, and generated inclusion and exclusion maps using segmentation centroids (to which slight Gaussian noise was added) to mimic user-generated points. We followed the official train/test split for ConSeP and used all extracted patches from PCNS for training. This resulted in 46,000 training and 8,600 validation patches. Training was performed for 60 epochs with a learning rate of 0.001, a batch size of 64 and Weighted BCE Dice loss, as described in the original paper, and we monitored validation performance using Dice score. Our best model achieved a Dice score of 0.874 in validation after 35 epochs, and was used for inference on collected pathologist annotations. Nuclick segmentations are then used as ground truth annotations for model training.

\subsection*{Vision Transformer}

The Vision Transformer (ViT) processes 2D images by first dividing the input into a sequence of flattened patches $(x_p)_{p \in \mathbb{R}^{N \times (P\times C)}}$, where $(H, W)$ is the original image resolution, $(P, P)$ is the patch size, and $N=H\times W/P^2$. Each patch is linearly projected into a $D$-dimensional embedding space, augmented with learnable 1D positional embeddings ($E_{\text{pos}}$) to retain spatial information. A [CLS] token prepended to the patch sequence serves as the global image representation for classification. The Transformer encoder consists of alternating multi-head self-attention \cite{ref50} and multi-layer perceptron (MLP) blocks, each preceded by a LayerNorm \cite{ref51} operation and followed by residual connections. The MLP employs two layers with Gaussian Error Linear Unit (GELU) activation. Unlike convolutional neural networks, ViT minimizes image-specific inductive biases. Spatial relations are learned entirely through self-attention, with locality restricted to the initial patch embedding and optional hybrid CNN feature inputs. The architecture is highly scalable, with common variants (Base, Large, Huge, Giant) differing in the number of layers, attention heads, and embedding dimension (see \cite{ref52} for detailed configurations).

\subsection*{CellViT}

\textbf{Model description}. CellViT \cite{ref40} adapts the UNETR architecture \cite{ref53} for 2D histopathology images by integrating a Vision Transformer \cite{ref42} encoder with a multi-task decoder structure inspired by HoVerNet. The model employs three distinct output branches: (1) a nuclei prediction (NP) branch for binary segmentation of nuclei boundaries, (2) a horizontal-vertical (HV) branch generating normalized distance maps to nuclei centers, and (3) a nuclei type (NT) branch for instance-aware classification. The ViT encoder processes image patches as token sequences with learnable positional embeddings, while skip connections fuse multi-scale features from five encoder stages into the decoder. Unlike traditional U-Nets, CellViT’s decoder uses isolated upsampling pathways for each branch, maintaining input resolution in the output. We selected different layers based on the size of the vision transformer encoder.
\\
\\
\textbf{Loss}. CellViT employs a multi-task loss combining weighted branch-specific objectives: (1) a nuclei prediction loss using cross-entropy and Dice to segment nuclei boundaries, (2) a horizontal-vertical loss with mean squared error on distance maps and their gradients for spatial localization and (3) a nuclei type loss with Focal Tversky to address class imbalance in fine-grained classification. The Focal Tversky loss prioritizes underrepresented nuclei classes through adjustable hyperparameters, while the total loss balances contributions via weights. We used the exact loss weights as presented in the original CellViT implementation (Table A3).
\\
\\
\textbf{Postprocessing}. To generate instance-aware segmentation from CellViT’s multi-branch outputs, we use the original two-stage approach, presented in HoVerNet. For individual patches, nuclei separation is achieved by: (1) computing gradients of the horizontal-vertical distance maps to detect boundary transitions, (2) applying Sobel edge detection to highlight nuclei contours, and (3) using marker-controlled watershed to resolve overlapping instances. Nuclei class assignments are derived via majority voting on the nuclei type prediction map within each segmented region.
\\
\\
\textbf{Handling multiple image sizes}. CellViT processes input images through a ViT backbone pretrained at a fixed resolution of 224x224, with positional encodings dynamically interpolated to accommodate varying input sizes. The encoder yields feature maps of spatial dimension 16x16 (for patch size 14) and 14x14 (for patch size 16), derived from the ratio of input resolution to patch size. During decoding, these feature maps are progressively upsampled (x24) to 256x256 (for patch size 14) or 224x224 (for patch size 16). For the 256x256 outputs, a final downscaling to 224x224 ensures consistency in loss computation and metric evaluation. This design preserves the positional information of the patches across scales while maintaining flexibility for multi-scale inputs.

\subsection*{Pathology Foundation Models}

\textbf{Self-supervised learning}. Self-supervised learning (SSL) enables models like ViTs to learn robust feature representations from unlabeled data by solving pretext tasks, eliminating the need for costly manual annotations. In computational pathology, SSL frameworks like iBOT \cite{ref26} (masked image modeling with online tokenization, where patches are reconstructed after random masking), DINO (self-distillation with no labels, using a teacher-student framework to match feature distributions across augmented views), and DINOv2 \cite{ref27} (a scalable extension of DINO with improved regularization, larger datasets, and efficient data pipelines) have become pivotal for pretrained foundation models. While iBOT focuses on local feature consistency through masked prediction, DINO and DINOv2 emphasize global representation alignment, with DINOv2 further optimizing stability and scalability for large-scale training. These methods leverage vision transformer architectures to capture hierarchical and semantically rich features from histopathology patches, which can later be fine-tuned for downstream tasks like cell detection, segmentation and classification. UNI and UNI2 were trained using DINOv2, enabling strong generalization across diverse tissue types. Virchow and Virchow2G were trained using DINOv2. Phikon and PhikonV2 were trained respectively using iBOT and DINOv2.
\\
\\
\textbf{Knowledge distillation}. Knowledge distillation in self-supervised learning (SSL) refers to the process of transferring knowledge from a large, pretrained teacher model to a smaller, more efficient student model without relying on labeled data. In this paradigm, the student learns by mimicking the teacher’s internal representations or similarity patterns across different views of the same image, enabling it to capture rich semantic features with far fewer parameters. A recent and compelling example is H0-mini, a distilled SSL model derived from a much larger foundation model (H‑Optimus‑0). H0-mini integrates techniques from DINO (global class token distillation) and iBOT (local patch-level supervision) to effectively compress the teacher’s knowledge. Despite its compact size (~86M parameters), H0-mini achieves state-of-the-art performance on benchmarks like HEST and EVA, demonstrating remarkable robustness across staining and scanner variations.

\subsection*{Evaluation metrics for detection, segmentation and classification}

\textbf{Detection and segmentation metrics}. To rigorously evaluate the segmentation performance of our model, we employed the panoptic quality (PQ) metric, which provides a comprehensive assessment by jointly considering detection accuracy and segmentation precision. The PQ score decomposes into two interpretable components: Detection Quality (DQ), analogous to the F1-score, which quantifies the correctness of instance detection, Segmentation Quality (SQ), defined as the average intersection-over-union (IoU) of matched segments.
Instance matching was performed using the Munkres (Hungarian) algorithm \cite{ref54, ref55}, an optimal assignment method that minimizes the total Euclidean distance between ground-truth and predicted centroids, ensuring one-to-one correspondence. Following established conventions \cite{ref56}, we consider a segment pair ($y$, $\hat{y}$) as a true positive (TP) only if their IoU exceeds 0.5, ensuring unique matching between predicted and ground-truth instances. Unmatched predictions and ground-truth segments are classified as false positives (FP) and false negatives (FN), respectively. We report the binary PQ, treating all nuclei as a single class. This approach aligns with recent benchmarks \cite{ref22, ref40}.
\\
\\
\textbf{Classification metrics}. Classification of nuclear type is performed on the instances extracted from either instance segmentation or detection. Thus, the evaluation of nuclear type classification must account for both detection and classification accuracy. For each nuclear type $t$, the detection task ($d$) partitions ground-truth (GT) and predicted instances into correctly detected instances ($\text{TP}_d$), missed GT instances ($\text{FN}_d$), and overdetected predictions ($\text{FP}_d$). The classification task ($c$) further divides $\text{TP}_d$ into correctly classified instances of type $t$ ($\text{TP}_c$), correctly classified instances of other types ($\text{TN}_c$), misclassified instances of type t ($\text{FP}_c$), and misclassified instances of other types ($\text{FN}_c$).
\\
\\
Classification metrics in the case of multi-label annotation. When the consensus yields several ground truth cell types, the previous approach cannot be used. Instead, for each nuclear type $t$, $\text{FP}_d$ is classically defined as the number of unpaired predicted cells whose prediction is $t$, while $\text{FN}_d$ is defined as the number of unpaired true cells whose ground-truth labels list contains $t$. For classification, we further define $\text{TP}_c$ if $t$ is in the ground-truth labels list when the prediction is $t$, $\text{TN}_c$ if $t$ is not in the ground-truth labels list and the prediction is not $t$, $\text{FP}_c$ if $t$ is not in the ground-truth labels list and the prediction is $t$ and $\text{FN}_c$ if both $t$ is in the ground-truth labels list, the prediction is $t$ and the prediction is not in ground-truth labels list.
\\
\\
Confidence intervals at 95\% confidence level were obtained by bootstrapping experiment results with 1000 repeats. All tests were two-tailed, and P-values $\le$ 0.05 were considered statistically significant.

\subsection*{External datasets}

\textbf{PanNuke}. The PanNuke dataset \cite{ref21} consists of 30,000 nuclei-annotated tiles split into 3 folds on H\&E images, coming from 19 different tissue types (Adrenal, Bile duct, Bladder, Breast, Colon, Cervix, Esophagus, Head and Neck, Kidney, Liver, Lung, Ovarian, Pancreatic, Prostate, Skin, Testis, Stomach, Thyroid, Uterus). Cells were annotated by trained pathologists on 5 different classes: Neoplastic, Non-Neoplastic Epithelial, Inflammatory, Connective and Dead. Each image has a dimension of 256x256 pixels at a 40x magnification.
\\
\\
\textbf{Lizard}. The Lizard dataset \cite{ref24} consists of 6 annotated datasets (DigestPath, CRAG, GlaS, PanNuke, CoNSeP, TCGA) for 495,179 annotated nuclei, coming from colon H\&E slides. Cells were annotated by trained pathologists on 6 different classes: Epithelial, Connective, Lymphocyte, Plasma, Neutrophil, Eosinophil. It is composed of 291 image regions with an average size of 1,016x917 pixels at a 20x magnification. We further pre-processed the raw images to create a preprocessed dataset consisting of images of size 224x224 pixels.

\subsection*{Benchmarking of CellViT on HistoTRAIN}

All models were trained using 3-fold cross-validation using the folds presented in \cite{ref21} on 4 Tesla T4 GPUs, with DeepSpeed Stage 0 (Data Parallelism) in mixed precision (FP16) and CPU offloading. Input images (40x magnification) of size 448x448 were directly fed to the CellViT model (see the Handling multiple image sizes paragraph for more details). We used Adam as an optimizer with a global batch size of 16 (4 per GPU), with an initial learning rate of 1e-5, weight decay of 1e-5 and betas (0.9, 0.99). We did not use any learning rate scheduler as we observed in our experiments that a low constant learning rate offered the lowest and steadiest decrease in validation loss. The models were trained for 150 epochs with validation every 10 epochs to reproduce the training setup of \cite{ref21}. We created skip connections that fuse features from encoder layers 3, 6, 9, and 12 (ViT-Base), layers 6, 12, 18 and 24 (ViT-Large), and layers 8, 16, 24 and 32 (ViT-Huge), ensuring consistent multi-scale fusion across model sizes. We used all the data augmentation parameters presented in the original CellViT implementation (Table A.2) \cite{ref40} and the same oversampling strategy to account for the cell distribution imbalance (paragraph 4.4 with $\gamma$ = 0.85). For full reproducibility, all experiments used a fixed random seed (111) for deterministic initialization and data shuffling.

\subsection*{External validation on HistoVAL}
To further assess the generalization ability of the cross-validated models on HistoTRAIN, the model from the last fold and last epoch was selected for external validation on HistoVAL.

\section{Data Availability}
TCGA data consists of whole-slide images and can be accessed through the NIH genomic data commons (\url{https://portal.gdc.cancer.gov}). The MOSAIC data is not publicly available but access to a subset of 60 patients can be requested through MOSAIC-Window (\url{https://www.mosaic-research.com/mosaic-window}). The PanNuke dataset is available at \url{https://huggingface.co/datasets/RationAI/PanNuke} and the Lizard dataset at \url{https://conic-challenge.grand-challenge.org/}.

\section{Code Availability}

To support broader TME biomarker research, we release the model weights and inference code at \url{https://github.com/owkin/histoplus/}, under a Creative Commons Attribution 4.0 International (CC-BY-4.0) License. An implementation of the NuClick model is available at \url{https://github.com/mostafajahanifar/nuclick_torch/}. The repositories referenced above contain license files and details of usage permissions. Any reuse of third-party software, including the NuClick implementation and the models used, complies with their respective licenses.

\section{Acknowledgements}
The present study was funded by Owkin. This study also makes use of data generated by the MOSAIC consortium (Owkin; Charité – Universitätsmedizin Berlin (DE); Lausanne University Hospital - CHUV (CH); Universitätsklinikum Erlangen (DE); Institut Gustave Roussy (FR); University of Pittsburgh (USA)). The authors thank Dr Kathrina Alexander, Dr Audrey Caudron, Dr Richard Doughty, Dr Romain Dubois, Dr Thibaut Gioanni, Dr Camelia Radulescu, Dr Thomas Rialland, Dr Pierre Romero and Dr Yannis Roxanis for their contributions to HistoTRAIN and HistoVAL. 

\putbib[article]
\end{bibunit}

\newpage
\appendix
\section*{Supplementary Material}
\FloatBarrier

\addcontentsline{toc}{section}{Supplementary Material}
\renewcommand{\tablename}{Supplementary Table}
\renewcommand{\figurename}{Supplementary Fig.}
\setcounter{table}{0}
\setcounter{figure}{0}
\setlength{\LTpre}{1pt} 
\setlength{\LTpost}{1pt}

\subsection*{}
\begin{table}[h!]
\centering
\setlength{\tabcolsep}{10pt} 
\renewcommand{\arraystretch}{1.2} 
\begin{tabular}{l*{3}{c}}
\toprule
\textbf{Encoder} & \textbf{Panoptic Quality} & \textbf{Detection Quality} & \textbf{Segmentation Quality} \\
\midrule
SAM-B     & 0.599 (0.011) & 0.725 (0.009) & 0.817 (0.004) \\
Phikon    & 0.606 (0.002) & 0.731 (0.003) & 0.823 (0.001) \\
Hibou-B   & 0.602 (0.018) & 0.722 (0.021) & 0.827 (0.004) \\
H0-mini   & 0.618 (0.003) & 0.742 (0.001) & 0.826 (0.005) \\
SAM-H     & 0.621 (0.014) & 0.738 (0.015) & 0.833 (0.004) \\
UNI 2     & 0.625 (0.006) & 0.747 (0.009) & 0.831 (0.002) \\
Virchow 2 & 0.633 (0.005) & 0.753 (0.007) & 0.835 (0.003) \\
\bottomrule
\end{tabular}
\caption{\textbf{Performance in cell detection and segmentation of encoders, using CellViT in cross-validation on HistoTRAIN ($n{=}1{,}415$)}. We report the mean metric values across folds, as well as the standard deviation in parentheses.}
\end{table}
\begin{sidewaystable}[p]
\centering
\setlength{\tabcolsep}{5pt}%
\renewcommand{\arraystretch}{1.2}%
\resizebox{\textheight}{!}{%
\begin{tabular}{l*{13}{c}}
\toprule
\textbf{Encoder} & \textbf{Cancer Cells} & \textbf{Lymphocytes} & \textbf{Fibroblasts} & \textbf{Plasmocytes} & \textbf{Eosinophils} & \textbf{Neutrophils} & \textbf{Macrophages} & \textbf{Smooth Muscle Cells} & \textbf{Endothelial Cells} & \textbf{Red Blood Cells} & \textbf{Epithelial Cells} & \textbf{Mitotic Figures} & \textbf{Apoptotic Body} \\
\midrule
SAM-B     & 0.567 (0.016) & 0.561 (0.018) & 0.344 (0.006) & 0.396 (0.015) & 0.476 (0.012) & 0.429 (0.015) & 0.158 (0.015) & 0.328 (0.044) & 0.207 (0.025) & 0.496 (0.059) & 0.260 (0.045) & 0.235 (0.064) & 0.214 (0.005) \\
Phikon    & 0.593 (0.047) & 0.573 (0.013) & 0.350 (0.006) & 0.443 (0.004) & 0.463 (0.036) & 0.449 (0.026) & 0.153 (0.005) & 0.469 (0.039) & 0.297 (0.016) & 0.538 (0.047) & 0.466 (0.061) & 0.275 (0.062) & 0.272 (0.007) \\
Hibou-B   & 0.601 (0.030) & 0.561 (0.013) & 0.362 (0.013) & 0.425 (0.013) & 0.487 (0.006) & 0.444 (0.025) & 0.131 (0.026) & 0.439 (0.023) & 0.271 (0.037) & 0.524 (0.021) & 0.441 (0.052) & 0.244 (0.071) & 0.245 (0.035) \\
H0-mini   & 0.621 (0.034) & 0.594 (0.011) & 0.380 (0.021) & 0.482 (0.013) & 0.511 (0.020) & 0.475 (0.024) & 0.172 (0.024) & 0.530 (0.043) & 0.340 (0.009) & 0.525 (0.076) & 0.466 (0.031) & 0.358 (0.077) & 0.254 (0.006) \\
SAM-H     & 0.589 (0.039) & 0.570 (0.025) & 0.366 (0.015) & 0.432 (0.010) & 0.484 (0.018) & 0.443 (0.022) & 0.147 (0.014) & 0.396 (0.044) & 0.231 (0.034) & 0.562 (0.035) & 0.359 (0.055) & 0.264 (0.048) & 0.257 (0.011) \\
UNI 2     & 0.626 (0.029) & 0.573 (0.022) & 0.371 (0.011) & 0.465 (0.023) & 0.499 (0.018) & 0.472 (0.028) & 0.140 (0.002) & 0.515 (0.039) & 0.344 (0.011) & 0.550 (0.047) & 0.521 (0.042) & 0.362 (0.077) & 0.259 (0.023) \\
Virchow 2 & 0.622 (0.025) & 0.571 (0.008) & 0.369 (0.023) & 0.457 (0.015) & 0.505 (0.014) & 0.462 (0.021) & 0.135 (0.016) & 0.481 (0.015) & 0.348 (0.014) & 0.530 (0.049) & 0.513 (0.051) & 0.332 (0.077) & 0.284 (0.003) \\
\bottomrule
\end{tabular}%
}
\caption{\textbf{Performance in cell classification of encoders, using CellViT in cross-validation on \textbf{HistoTRAIN} ($n{=}1{,}415$)}. We report the mean across folds with the standard deviation in parentheses.}
\label{tab:suppl_cell_classification}
\end{sidewaystable}

\begin{table}[h!]
\centering
\setlength{\tabcolsep}{10pt} 
\renewcommand{\arraystretch}{1.3} 
\label{tab:encoder_performance}
\begin{tabular}{l*{3}{c}}
\toprule
\textbf{Encoder} & \textbf{Panoptic Quality} & \textbf{Detection Quality} & \textbf{Segmentation Quality} \\
\midrule
SAM-B & 0.599 [0.590 - 0.608] & 0.753 [0.743 - 0.764] & 0.792 [0.790 - 0.795] \\
Phikon & 0.583 [0.574 - 0.592] & 0.744 [0.734 - 0.753] & 0.781 [0.778 - 0.784] \\
Hibou-B & 0.588 [0.579 - 0.597] & 0.741 [0.730 - 0.751] & 0.792 [0.789 - 0.795] \\
H0-mini & 0.605 [0.595 - 0.613] & 0.753 [0.742 - 0.763] & 0.801 [0.799 - 0.804] \\
SAM-H & 0.581 [0.570 - 0.591] & 0.716 [0.703 - 0.728] & 0.808 [0.805 - 0.810] \\
UNI 2 & 0.597 [0.587 - 0.606] & 0.753 [0.742 - 0.763] & 0.791 [0.788 - 0.794] \\
Virchow 2 & 0.596 [0.586 - 0.604] & 0.753 [0.742 - 0.763] & 0.790 [0.787 - 0.792] \\
\bottomrule
\end{tabular}
\caption{\textbf{Performance of encoders in cell detection and segmentation, using CellViT in external validation on HistoVAL ($n{=}530$)}. Median and confidence intervals at 95\% confidence level were obtained by bootstrapping experiment results with 1000 repeats.}
\end{table}
\begin{sidewaystable}[p]
\centering
\setlength{\tabcolsep}{5pt}
\renewcommand{\arraystretch}{1.2}
\resizebox{\textheight}{!}{%
\begin{tabular}{l*{13}{c}}
\toprule
\textbf{Encoder} &
\textbf{Cancer Cells} &
\textbf{Lymphocytes} &
\textbf{Fibroblasts} &
\textbf{Plasmocytes} &
\textbf{Eosinophils} &
\textbf{Neutrophils} &
\textbf{Macrophages} &
\textbf{Smooth Muscle Cells} &
\textbf{Endothelial Cells} &
\textbf{Red Blood Cells} &
\textbf{Epithelial Cells} &
\textbf{Mitotic Figures} &
\textbf{Apoptotic Body} \\
\midrule
SAM-B &
\makecell{0.456\\\footnotesize[0.414-0.493]} &
\makecell{0.558\\\footnotesize[0.515-0.601]} &
\makecell{0.325\\\footnotesize[0.300-0.345]} &
\makecell{0.269\\\footnotesize[0.219-0.322]} &
\makecell{0.386\\\footnotesize[0.302-0.453]} &
\makecell{0.295\\\footnotesize[0.232-0.348]} &
\makecell{0.163\\\footnotesize[0.140-0.185]} &
\makecell{0.186\\\footnotesize[0.110-0.257]} &
\makecell{0.207\\\footnotesize[0.174-0.239]} &
\makecell{0.466\\\footnotesize[0.377-0.555]} &
\makecell{0.023\\\footnotesize[0.005-0.049]} &
\makecell{0.205\\\footnotesize[0.097-0.324]} &
\makecell{0.143\\\footnotesize[0.122-0.167]} \\
Phikon &
\makecell{0.547\\\footnotesize[0.510-0.583]} &
\makecell{0.668\\\footnotesize[0.632-0.697]} &
\makecell{0.357\\\footnotesize[0.334-0.380]} &
\makecell{0.430\\\footnotesize[0.380-0.476]} &
\makecell{0.300\\\footnotesize[0.221-0.385]} &
\makecell{0.196\\\footnotesize[0.144-0.258]} &
\makecell{0.151\\\footnotesize[0.131-0.173]} &
\makecell{0.279\\\footnotesize[0.180-0.390]} &
\makecell{0.301\\\footnotesize[0.260-0.340]} &
\makecell{0.492\\\footnotesize[0.430-0.549]} &
\makecell{0.371\\\footnotesize[0.260-0.470]} &
\makecell{0.100\\\footnotesize[0.039-0.177]} &
\makecell{0.096\\\footnotesize[0.076-0.119]} \\
Hibou-B &
\makecell{0.527\\\footnotesize[0.493-0.561]} &
\makecell{0.648\\\footnotesize[0.611-0.680]} &
\makecell{0.374\\\footnotesize[0.348-0.397]} &
\makecell{0.403\\\footnotesize[0.341-0.456]} &
\makecell{0.418\\\footnotesize[0.331-0.486]} &
\makecell{0.249\\\footnotesize[0.196-0.295]} &
\makecell{0.169\\\footnotesize[0.145-0.192]} &
\makecell{0.323\\\footnotesize[0.198-0.456]} &
\makecell{0.253\\\footnotesize[0.210-0.302]} &
\makecell{0.483\\\footnotesize[0.401-0.552]} &
\makecell{0.223\\\footnotesize[0.102-0.349]} &
\makecell{0.129\\\footnotesize[0.021-0.281]} &
\makecell{0.115\\\footnotesize[0.094-0.135]} \\
H0-mini &
\makecell{0.583\\\footnotesize[0.544-0.619]} &
\makecell{0.688\\\footnotesize[0.656-0.716]} &
\makecell{0.377\\\footnotesize[0.353-0.400]} &
\makecell{0.483\\\footnotesize[0.433-0.529]} &
\makecell{0.456\\\footnotesize[0.372-0.520]} &
\makecell{0.245\\\footnotesize[0.181-0.315]} &
\makecell{0.219\\\footnotesize[0.187-0.250]} &
\makecell{0.403\\\footnotesize[0.282-0.515]} &
\makecell{0.333\\\footnotesize[0.292-0.378]} &
\makecell{0.504\\\footnotesize[0.424-0.571]} &
\makecell{0.422\\\footnotesize[0.279-0.536]} &
\makecell{0.245\\\footnotesize[0.181-0.315]} &
\makecell{0.125\\\footnotesize[0.098-0.153]} \\
SAM-H &
\makecell{0.531\\\footnotesize[0.491-0.567]} &
\makecell{0.653\\\footnotesize[0.614-0.685]} &
\makecell{0.386\\\footnotesize[0.359-0.408]} &
\makecell{0.387\\\footnotesize[0.337-0.435]} &
\makecell{0.391\\\footnotesize[0.300-0.462]} &
\makecell{0.247\\\footnotesize[0.181-0.313]} &
\makecell{0.205\\\footnotesize[0.182-0.230]} &
\makecell{0.245\\\footnotesize[0.125-0.369]} &
\makecell{0.212\\\footnotesize[0.175-0.256]} &
\makecell{0.501\\\footnotesize[0.410-0.577]} &
\makecell{0.121\\\footnotesize[0.053-0.207]} &
\makecell{0.093\\\footnotesize[0.024-0.194]} &
\makecell{0.092\\\footnotesize[0.068-0.119]} \\
UNI 2 &
\makecell{0.603\\\footnotesize[0.570-0.634]} &
\makecell{0.705\\\footnotesize[0.676-0.730]} &
\makecell{0.401\\\footnotesize[0.377-0.425]} &
\makecell{0.507\\\footnotesize[0.455-0.553]} &
\makecell{0.474\\\footnotesize[0.391-0.535]} &
\makecell{0.292\\\footnotesize[0.239-0.332]} &
\makecell{0.168\\\footnotesize[0.141-0.194]} &
\makecell{0.310\\\footnotesize[0.217-0.412]} &
\makecell{0.332\\\footnotesize[0.286-0.379]} &
\makecell{0.471\\\footnotesize[0.391-0.542]} &
\makecell{0.570\\\footnotesize[0.437-0.671]} &
\makecell{0.228\\\footnotesize[0.112-0.360]} &
\makecell{0.087\\\footnotesize[0.066-0.112]} \\
Virchow 2 &
\makecell{0.579\\\footnotesize[0.544-0.613]} &
\makecell{0.675\\\footnotesize[0.643-0.702]} &
\makecell{0.362\\\footnotesize[0.339-0.385]} &
\makecell{0.459\\\footnotesize[0.407-0.510]} &
\makecell{0.409\\\footnotesize[0.321-0.484]} &
\makecell{0.250\\\footnotesize[0.205-0.299]} &
\makecell{0.202\\\footnotesize[0.176-0.229]} &
\makecell{0.341\\\footnotesize[0.243-0.437]} &
\makecell{0.348\\\footnotesize[0.303-0.394]} &
\makecell{0.447\\\footnotesize[0.370-0.517]} &
\makecell{0.413\\\footnotesize[0.273-0.542]} &
\makecell{0.207\\\footnotesize[0.095-0.326]} &
\makecell{0.080\\\footnotesize[0.062-0.103]} \\
\bottomrule
\end{tabular}%
}
\caption{\textbf{Performance of encoders in cell classification, using CellViT in external validation on \textbf{HistoVAL} ($n{=}530$)}. Medians and 95\% confidence intervals were obtained by bootstrapping with 1000 repeats.}
\label{tab:suppl_external_cls}
\end{sidewaystable}

\begin{table}[h!]
\centering
\setlength{\tabcolsep}{10pt} 
\renewcommand{\arraystretch}{1.3} 
\begin{tabular}{l*{3}{c}}
\toprule
\textbf{Architecture} & \textbf{Encoder} & \textbf{Number of parameters} & \textbf{Memory requirements (in GB)} \\
\midrule
CellViT & ViT-Base & 146,090,259 & 14.2 \\
CellViT & ViT-Huge & 748,430,355 & 28.7 \\
\bottomrule
\end{tabular}
\caption{\textbf{Computational cost for CellViT with a vision transformer base (Phikon) and huge (UNI 2) as encoders.} To have a common ground of comparison between encoders, the memory requirements were evaluated during training in FP32, on the same machine with 4xTesla T4 GPUs (16GB of RAM each), using DeepSpeed ZeRO.}
\end{table}

\begin{table}[h!]
\centering
\setlength{\tabcolsep}{12pt} 
\renewcommand{\arraystretch}{1.3} 
\begin{tabular}{l*{3}{c}}
\toprule
\textbf{Indication} & \textbf{Panoptic Quality} & \textbf{Detection Quality} & \textbf{Segmentation Quality} \\
\midrule
Bladder       & 0.526 [0.495 - 0.555] & 0.656 [0.620 - 0.687] & 0.799 [0.790 - 0.808] \\
Mesothelioma  & 0.598 [0.576 - 0.617] & 0.742 [0.718 - 0.766] & 0.804 [0.798 - 0.809] \\
Lung          & 0.617 [0.598 - 0.632] & 0.756 [0.737 - 0.774] & 0.814 [0.809 - 0.819] \\
Colon         & 0.581 [0.564 - 0.597] & 0.736 [0.717 - 0.754] & 0.788 [0.782 - 0.794] \\
Ovarian       & 0.647 [0.626 - 0.667] & 0.805 [0.782 - 0.825] & 0.803 [0.795 - 0.810] \\
Breast        & 0.671 [0.656 - 0.686] & 0.836 [0.819 - 0.852] & 0.801 [0.796 - 0.807] \\
\bottomrule
\end{tabular}
\caption{\textbf{Performance of HistoPLUS in cell detection and segmentation in external validation on HistoVAL, stratified by indication.} Mean and confidence intervals at 95\% confidence level were obtained by bootstrapping experiment results with 1000 repeats.}
\end{table}

\begin{sidewaystable}[p]
\centering
\setlength{\tabcolsep}{5pt}
\renewcommand{\arraystretch}{1.2}
\resizebox{\textheight}{!}{%
\begin{tabular}{l*{13}{c}}
\toprule
\textbf{F1 scores} &
\textbf{Cancer Cells} &
\textbf{Lymphocytes} &
\textbf{Fibroblasts} &
\textbf{Plasmocytes} &
\textbf{Eosinophils} &
\textbf{Neutrophils} &
\textbf{Macrophages} &
\textbf{Smooth Muscle Cells} &
\textbf{Endothelial Cells} &
\textbf{Red Blood Cells} &
\textbf{Epithelial Cells} &
\textbf{Mitotic Figures} &
\textbf{Apoptotic Body} \\
\midrule
Bladder &
\makecell{0.620\\\footnotesize[0.548-0.679]} &
\makecell{0.470\\\footnotesize[0.365-0.538]} &
\makecell{0.368\\\footnotesize[0.294-0.445]} &
\makecell{0.332\\\footnotesize[0.162-0.486]} &
\makecell{0.500\\\footnotesize[0.229-0.722]} &
\makecell{0.178\\\footnotesize[0.088-0.292]} &
\makecell{0.122\\\footnotesize[0.042-0.209]} &
\makecell{0.523\\\footnotesize[0.345-0.656]} &
\makecell{0.436\\\footnotesize[0.290-0.580]} &
\makecell{0.415\\\footnotesize[0.245-0.566]} &
\makecell{N/A} &
\makecell{0.750\\\footnotesize[0.667-1.000]} &
\makecell{0.211\\\footnotesize[0.123-0.318]} \\
Meso &
\makecell{0.677\\\footnotesize[0.642-0.710]} &
\makecell{0.557\\\footnotesize[0.490-0.615]} &
\makecell{0.333\\\footnotesize[0.294-0.371]} &
\makecell{0.605\\\footnotesize[0.395-0.716]} &
\makecell{0.465\\\footnotesize[0.087-0.648]} &
\makecell{0.106\\\footnotesize[0.045-0.183]} &
\makecell{0.128\\\footnotesize[0.061-0.196]} &
\makecell{0.349\\\footnotesize[0.040-0.565]} &
\makecell{0.345\\\footnotesize[0.235-0.470]} &
\makecell{0.463\\\footnotesize[0.324-0.608]} &
\makecell{0.199\\\footnotesize[0.000-0.352]} &
\makecell{N/A} &
\makecell{0.166\\\footnotesize[0.124-0.206]} \\
Lung &
\makecell{0.538\\\footnotesize[0.426-0.631]} &
\makecell{0.626\\\footnotesize[0.573-0.666]} &
\makecell{0.310\\\footnotesize[0.262-0.361]} &
\makecell{0.588\\\footnotesize[0.504-0.632]} &
\makecell{0.517\\\footnotesize[0.386-0.673]} &
\makecell{0.223\\\footnotesize[0.139-0.343]} &
\makecell{0.241\\\footnotesize[0.191-0.291]} &
\makecell{N/A} &
\makecell{0.352\\\footnotesize[0.258-0.440]} &
\makecell{0.700\\\footnotesize[0.616-0.749]} &
\makecell{0.058\\\footnotesize[0.000-0.156]} &
\makecell{0.143\\\footnotesize[0.000-0.667]} &
\makecell{0.088\\\footnotesize[0.043-0.151]} \\
Colon &
\makecell{0.427\\\footnotesize[0.319-0.533]} &
\makecell{0.554\\\footnotesize[0.498-0.603]} &
\makecell{0.353\\\footnotesize[0.306-0.402]} &
\makecell{0.431\\\footnotesize[0.362-0.495]} &
\makecell{0.454\\\footnotesize[0.346-0.534]} &
\makecell{0.290\\\footnotesize[0.189-0.384]} &
\makecell{0.135\\\footnotesize[0.106-0.169]} &
\makecell{0.255\\\footnotesize[0.098-0.460]} &
\makecell{0.160\\\footnotesize[0.092-0.228]} &
\makecell{0.461\\\footnotesize[0.274-0.577]} &
\makecell{0.534\\\footnotesize[0.349-0.653]} &
\makecell{0.118\\\footnotesize[0.000-0.500]} &
\makecell{0.108\\\footnotesize[0.066-0.168]} \\
Ovarian &
\makecell{0.682\\\footnotesize[0.594-0.751]} &
\makecell{0.639\\\footnotesize[0.585-0.676]} &
\makecell{0.474\\\footnotesize[0.417-0.526]} &
\makecell{0.330\\\footnotesize[0.183-0.430]} &
\makecell{0.250\\\footnotesize[0.000-0.889]} &
\makecell{0.112\\\footnotesize[0.000-0.261]} &
\makecell{0.182\\\footnotesize[0.084-0.278]} &
\makecell{N/A} &
\makecell{0.268\\\footnotesize[0.133-0.411]} &
\makecell{0.354\\\footnotesize[0.148-0.531]} &
\makecell{N/A} &
\makecell{N/A} &
\makecell{0.131\\\footnotesize[0.067-0.170]} \\
Breast &
\makecell{0.475\\\footnotesize[0.302-0.620]} &
\makecell{0.799\\\footnotesize[0.757-0.829]} &
\makecell{0.463\\\footnotesize[0.409-0.523]} &
\makecell{0.296\\\footnotesize[0.120-0.380]} &
\makecell{0.429\\\footnotesize[0.100-0.727]} &
\makecell{0.139\\\footnotesize[0.033-0.245]} &
\makecell{0.337\\\footnotesize[0.270-0.404]} &
\makecell{0.409\\\footnotesize[0.000-0.580]} &
\makecell{0.394\\\footnotesize[0.325-0.472]} &
\makecell{0.382\\\footnotesize[0.225-0.486]} &
\makecell{0.466\\\footnotesize[0.000-0.745]} &
\makecell{0.235\\\footnotesize[0.000-0.429]} &
\makecell{0.089\\\footnotesize[0.020-0.159]} \\
\bottomrule
\end{tabular}%
}
\caption{\textbf{Performance of HistoPLUS in cell classification in external validation on HistoVAL, stratified by indication}. Means and 95\% confidence intervals were obtained by bootstrapping experiment results with 1000 repeats.}
\label{tab:suppl_histoplus_by_indication}
\end{sidewaystable}
\begin{table}[p]
\centering
\setlength{\tabcolsep}{8pt}
\renewcommand{\arraystretch}{1.3}
\begin{tabular}{llccc}
\toprule
 & \textbf{Model} & \textbf{Panoptic Quality} & \textbf{Detection Quality} & \textbf{Segmentation Quality} \\
\midrule
\multirow{2}{*}{\textbf{Seen at training time}} 
& CellViT SAM-H & \makecell{0.559\\\footnotesize[0.547--0.571]} & \makecell{0.688\\\footnotesize[0.674--0.703]} & \makecell{0.808\\\footnotesize[0.805--0.811]} \\
& HistoPLUS & \makecell{0.582\\\footnotesize[0.572--0.594]} & \makecell{0.725\\\footnotesize[0.713--0.738]} & \makecell{0.801\\\footnotesize[0.798--0.804]} \\
\midrule
\multirow{2}{*}{\textbf{Unseen at training time}} 
& CellViT SAM-H & \makecell{0.637\\\footnotesize[0.622--0.651]} & \makecell{0.788\\\footnotesize[0.772--0.806]} & \makecell{0.807\\\footnotesize[0.802--0.811]} \\
& HistoPLUS & \makecell{0.663\\\footnotesize[0.650--0.675]} & \makecell{0.826\\\footnotesize[0.812--0.840]} & \makecell{0.802\\\footnotesize[0.797--0.807]} \\
\bottomrule
\end{tabular}
\caption{\textbf{Comparison of performance of HistoPLUS and CellViT SAM-H in cell detection and segmentation on indications seen vs. unseen at training time (external validation on HistoVAL)}. Means and 95\% confidence intervals were obtained by bootstrapping experiment results with 1000 repeats.}
\label{tab:suppl_histoplus_vs_samh}
\end{table}

\begin{sidewaystable}[p]
\centering
\setlength{\tabcolsep}{3.6pt}
\scriptsize
\resizebox{!}{0.06\textheight}{%
\begin{tabular}{ll*{13}{c}}
\toprule
 & \textbf{Model} &
 \textbf{Cancer Cells} &
 \textbf{Lymphocytes} &
 \textbf{Fibroblasts} &
 \textbf{Plasmocytes} &
 \textbf{Eosinophils} &
 \textbf{Neutrophils} &
 \textbf{Macrophages} &
 \makecell{\textbf{Smooth}\\\textbf{Muscle Cells}} &
 \makecell{\textbf{Endothelial}\\\textbf{Cells}} &
 \makecell{\textbf{Red Blood}\\\textbf{Cells}} &
 \textbf{Epithelial Cells} &
 \textbf{Mitotic Figures} &
 \textbf{Apoptotic Body} \\
\midrule
\multirow{2}{*}{\textbf{Seen at training time}}
& CellViT SAM-H
& \makecell{0.525\\\footnotesize[0.479--0.566]}
& \makecell{0.528\\\footnotesize[0.490--0.563]}
& \makecell{0.349\\\footnotesize[0.323--0.375]}
& \makecell{0.419\\\footnotesize[0.364--0.470]}
& \makecell{0.400\\\footnotesize[0.303--0.486]}
& \makecell{0.247\\\footnotesize[0.180--0.317]}
& \makecell{0.169\\\footnotesize[0.141--0.197]}
& \makecell{0.265\\\footnotesize[0.136--0.392]}
& \makecell{0.229\\\footnotesize[0.181--0.281]}
& \makecell{0.532\\\footnotesize[0.433--0.613]}
& \makecell{0.111\\\footnotesize[0.043--0.188]}
& \makecell{0.102\\\footnotesize[0.000--0.235]}
& \makecell{0.096\\\footnotesize[0.068--0.129]} \\
& HistoPLUS
& \makecell{0.582\\\footnotesize[0.539--0.620]}
& \makecell{0.568\\\footnotesize[0.534--0.596]}
& \makecell{0.339\\\footnotesize[0.313--0.366]}
& \makecell{0.510\\\footnotesize[0.453--0.557]}
& \makecell{0.464\\\footnotesize[0.373--0.535]}
& \makecell{0.255\\\footnotesize[0.183--0.327]}
& \makecell{0.180\\\footnotesize[0.152--0.207]}
& \makecell{0.412\\\footnotesize[0.282--0.526]}
& \makecell{0.312\\\footnotesize[0.258--0.364]}
& \makecell{0.534\\\footnotesize[0.442--0.606]}
& \makecell{0.429\\\footnotesize[0.275--0.545]}
& \makecell{0.200\\\footnotesize[0.042--0.377]}
& \makecell{0.125\\\footnotesize[0.099--0.162]} \\
\midrule
\multirow{2}{*}{\textbf{Unseen at training time}}
& CellViT SAM-H
& \makecell{0.553\\\footnotesize[0.462--0.630]}
& \makecell{0.734\\\footnotesize[0.687--0.777]}
& \makecell{0.471\\\footnotesize[0.427--0.512]}
& \makecell{0.213\\\footnotesize[0.127--0.294]}
& \makecell{0.267\\\footnotesize[0.122--0.588]}
& \makecell{0.301\\\footnotesize[0.066--0.421]}
& \makecell{0.272\\\footnotesize[0.229--0.317]}
& \makecell{0.044\\\footnotesize[0.000--0.111]}
& \makecell{0.189\\\footnotesize[0.132--0.250]}
& \makecell{0.350\\\footnotesize[0.199--0.465]}
& \makecell{0.184\\\footnotesize[0.000--0.425]}
& \makecell{0.074\\\footnotesize[0.000--0.286]}
& \makecell{0.069\\\footnotesize[0.032--0.105]} \\
& HistoPLUS
& \makecell{0.589\\\footnotesize[0.493--0.668]}
& \makecell{0.775\\\footnotesize[0.737--0.805]}
& \makecell{0.471\\\footnotesize[0.426--0.510]}
& \makecell{0.321\\\footnotesize[0.211--0.393]}
& \makecell{0.343\\\footnotesize[0.163--0.667]}
& \makecell{0.122\\\footnotesize[0.056--0.212]}
& \makecell{0.298\\\footnotesize[0.241--0.358]}
& \makecell{0.346\\\footnotesize[0.000--0.520]}
& \makecell{0.363\\\footnotesize[0.297--0.428]}
& \makecell{0.377\\\footnotesize[0.244--0.482]}
& \makecell{0.390\\\footnotesize[0.000--0.682]}
& \makecell{0.172\\\footnotesize[0.000--0.300]}
& \makecell{0.117\\\footnotesize[0.067--0.156]} \\
\bottomrule
\end{tabular}%
}
\caption{\textbf{Performance in cell classification of HistoPLUS and CellViT SAM-H on HistoVAL, split by indications seen vs.\ unseen at training time}. Means and 95\% confidence intervals obtained by bootstrapping with 1000 repeats.}
\label{tab:suppl_table_9}
\end{sidewaystable}

\clearpage
\FloatBarrier 

\begin{figure}[p]
  \centering
  \includegraphics[width=0.9\textwidth]{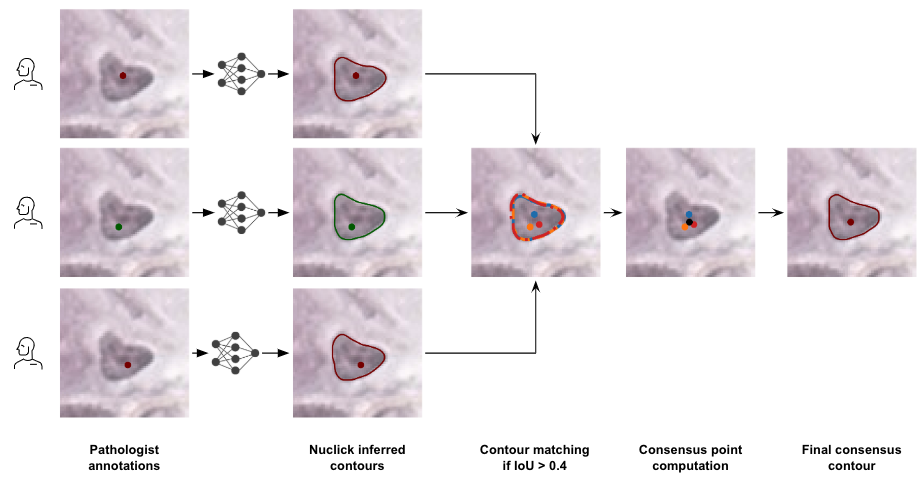}
  \caption*{\textbf{Supplementary Fig.1.} 
Consensus annotations are derived from multiple expert pathologists’ inputs following multiple steps: 
1) Multiple pathologists annotate nuclei centroids on tiles, 
2) NuClick expands points to nuclei contours, 
3) annotations are matched (IoU $>$ 0.4), retaining nuclei identified by at least two pathologists, 
4) consensus centroids are computed from the pathologist-annotated centroids, cell types are determined by majority voting, and final contours are generated by inferring NuClick on consensus points.}
  \label{fig:supp_last}
\end{figure}

\begin{figure}[p]
  \centering
  \includegraphics[width=0.9\textwidth]{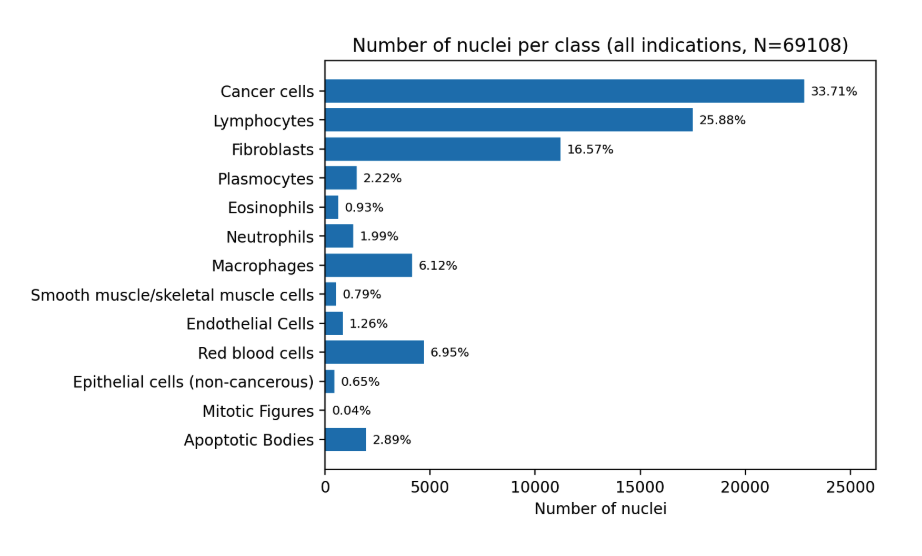}
  \caption*{\textbf{Supplementary Fig.2.} Resulting number of nuclei per class for all indications in our external validation set (N=69,108), after consensus annotations.}
  \label{fig:supp_last}
\end{figure}

\clearpage
\end{document}